\documentclass[10pt, a4paper]{article}

\usepackage[final]{lrec2026} 

\usepackage{comment}
\usepackage{graphicx}
\usepackage{url}
\usepackage{tabularx}
\usepackage{inconsolata}
\usepackage{microtype}
\usepackage{latexsym}
\usepackage{longtable}
\usepackage{booktabs}
\usepackage{subcaption}
\usepackage{amsmath}
\usepackage{supertabular}
\usepackage{booktabs}
\usepackage{threeparttable}

\title{\textit{Floating} or \textit{Suggesting} \textit{Ideas}? A Large-Scale Contrastive Analysis of Metaphorical and Literal Verb–Object Constructions}

\name{Prisca Piccirilli\textsuperscript{1} {\normalfont and} Alexander Fraser\textsuperscript{2,3} {\normalfont and} Sabine Schulte im Walde\textsuperscript{1}} 

\address{\textsuperscript{1}Institute for Natural Language Processing ({\small IMS}), University of Stuttgart, Germany \\
  \textsuperscript{2}Technical University of Munich ({\small TUM}), Germany \\
  \textsuperscript{3}Munich Center for Machine Learning ({\small MCML}), Germany \\
         \{prisca.piccirilli, schulte\}@ims.uni-stuttgart.de\\}

\abstract{
Metaphor is a pervasive feature of everyday language, enabling speakers to express
abstract concepts in terms of more concrete domains. 
While prior work has extensively examined metaphors from cognitive and psycholinguistic perspectives, large-scale empirical comparisons between metaphorical and literal language remain limited, especially when they compete in conveying the same meaning. 
We address this gap through a systematic contrastive analysis of 297 near-synonymous English verb–object ({\small VO}) pairs (e.g., \textit{float idea} vs. \textit{suggest idea}) 
in 
approximately 2 million extracted corpus sentences, allowing us to examine their contextual usage.
Using five established {\small NLP} tools -- {\small SEANCE, TAALES, TAASSC, TAACO} and TAALED -- we derive 2,293 metaphor-related cognitive and linguistic features capturing affective, lexical, syntactic, and discourse-level properties.
We address two research questions: \\(i) whether these features consistently differ between metaphorical and literal language (\textbf{cross-pair analysis}), and (ii) whether individual {\small VO} pairs show strong internal divergence between their metaphorical and literal variants (\textbf{within-pair analysis}). 
Cross-pair results reveal robust global tendencies: literal contexts are associated with higher lexical frequency, stronger cohesion, and greater structural regularity, whereas metaphorical contexts show increased affective loading, imageability, lexical diversity, and constructional specificity. In contrast, within-pair analyses reveal substantial heterogeneity and most pairs display non-uniform directional effects across cognitive or linguistic dimensions.
Our findings suggest that there is no single, consistent distributional pattern that consistently distinguishes metaphorical from literal language. Instead, the differences we observe are largely construction-specific. 
Overall, by combining large-scale data with a broad range of cognitive and linguistic features, this study offers a fine-grained understanding of the contrast between metaphorical and literal {\small VO} usages.
 \\ \newline  }

\begin{document}

\maketitleabstract

\section{Introduction}

\begin{figure*}[h!]
    \centering
    \includegraphics[width=\textwidth]{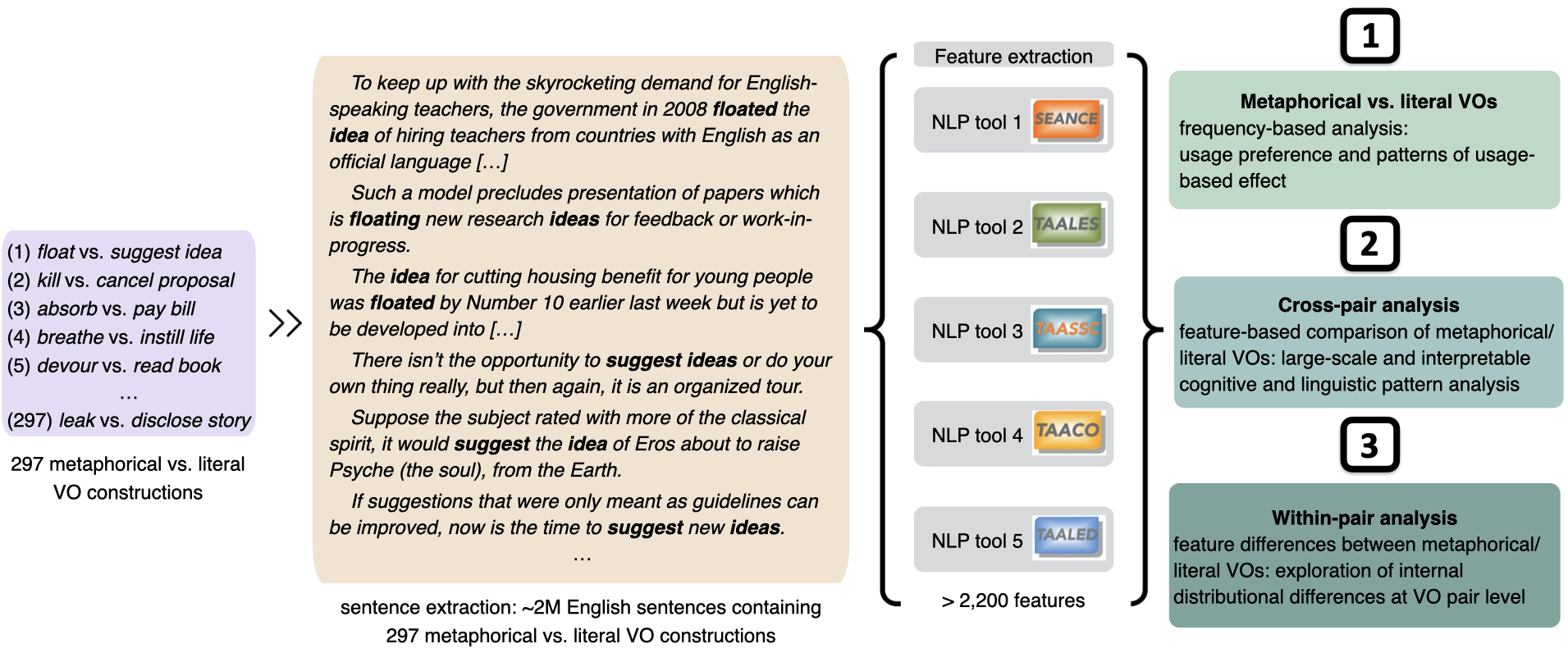}
    \caption{Pipeline of our approach: based on an existing set of metaphorival vs. literal {\small VO} pairs, we extract sentences containing these {\small VO}s. We then run five {\small NLP} tools on all sentences (masking the {\small VO}, see Section \ref{sec:approach}) to extract more than 2,200 cognitive and linguistic features in order to analyze the contrast between metaphorical and literal language usage from (i) a cross-pair and (ii) a within-pair perspective.}
    \label{fig:pipeline}
\end{figure*}

Metaphor is a "necessary" language feature of everyday thought and communication that allows speakers to conceptualize abstract 
ideas in terms of more concrete domains \citep[i.a.]{Ortony:1975, Lakoff-Johnson:1980, vandenBroek:1981, Schaeffner:2004}. 
While metaphor has been extensively studied in cognitive linguistics and psycholinguistics \citep[i.a.]{Gibbs:1989, Blasko:99, Giora:2002, Glucksberg:03, Steen:2004}, its empirical analysis in contrast to literal language use remains limited. 
For instance, \textit{float the idea} and \textit{suggest the idea} both convey the act of proposing an idea, but they differ in register, connotation, and metaphorical framing: 
\textit{float the idea} employs a conceptual metaphor ({\small IDEAS ARE OBJECTS IN MOTION}), projecting an abstract act into a concrete physical domain \citep{Lakoff-Johnson:1980}. In contrast, \textit{suggest the idea} is literal and denotes the speech act directly.
Pragmatically, the verb \textit{float} conveys tentativeness or strategic intent, while \textit{suggest} is more neutral and direct, and commonly used in formal and informal contexts \citep{Cameron-et-al:2003, Shutova:2010}.

Understanding how metaphorical and literal expressions differ at scale is crucial not only for theoretical insights but also for practical applications such as machine translation, sentiment analysis, and information retrieval. 
In this paper, we fill this gap by conducting a large-scale, systematic analysis of metaphorical and literal language in contrast.
We curate and analyze a dataset of approximately 2 million sentences containing semantically equivalent verb-object ({\small VO}) pairs, where one variant is metaphorical, e.g., \textit{float idea}, and the other is literal, e.g., \textit{suggest idea}. This paired design enables controlled comparison across usage contexts while holding the underlying meaning constant.
Using a variety
of {\small NLP} tools\footnote{\url{https://www.linguisticanalysistools.org/}}, we automatically extract a wide range of over 2,200 features, allowing us to explore cognitive, distributional and structural differences between metaphorical and literal usages. 
We conduct a large-scale and thorough descriptive analysis of these contrasts along two axes: 
we identify (i)~which cognitive and linguistic features systematically distinguish metaphorical from literal language across VO pairs (\textbf{cross-pair}), and (ii)~which {\small VO} pairs are “most strongly metaphorical” or “most strongly literal” across features but \textbf{within-pair}.
Figure~\ref{fig:pipeline} provides a visual pipeline of our work study.
%
In sum, our contributions are threefold:
\begin{itemize}
    \item We present a large-scale, contrastive dataset of metaphorical and literal verb-object expressions in naturalistic English text.
    \item We conduct an empirical analysis of contextual feature-based differences between metaphorical and literal verb-object expressions.
    \item We shed light on the internal metaphorical-literal differences at the verb-object level.  
\end{itemize}
By focusing on metaphorical and literal paraphrases with shared
meaning, this work advances the computational study of figurative language beyond detection, offering a scalable and interpretable approach to metaphor-literal contrast. Our findings open new avenues for metaphor modeling and the injection of cognitive and linguistic knowledge, and it underlines the importance of metaphorical variation in language understanding tasks\footnote{All data and analysis files, including the full set of detailed outputs, are available in our repository at \url{https://github.com/priscapiccirilli/Met-Lit-Contrast}.}. 

\section{Datasets and Linguistic Features}
\label{sec:data}
\paragraph{{\small VO} Pairs} We
downloaded the {\small VO}s from \citet{Piccirilli-et-al:2024}. They collected a set of 47 {\small VO}s from previous work \citep{Mohammad-etal:2016, Shutova:2010, Piccirilli-SchulteImWalde:2021, Stowe-et-al:2022}, which they semi-automatically extended by collecting the most frequently observed shared direct objects of the corresponding verbs, resulting in a total of 297 metaphorical {\small VO}s and their corresponding literal paraphrases.

\paragraph{{\small VO} Sentences} 
\citet{Piccirilli-et-al:2024} extracted 1,691 English sentences containing this set of {\small VO} pairs.
We thus use this publicly available dataset, {\small VOLIMET}\footnote{\url{https://github.com/priscapiccirilli/VOLIMET}}.
However, since our goal is to analyze large amounts of natural language,
we enlarge this dataset by automatically extracting all available English sentences containing the {\small VO}s from the {\small ENCOW} corpus \citep[{\small EN}glish {\small CO}rpora from the {\small W}eb,][]{Schafer-Bildhauer:2012, Schafer:2015}, where the object is the direct object of the verb. 
We extract the entire sentence regardless of its length or the distance between the verb and object, which allows us to capture contextual variations that are directly relevant to our features. 
{\small ENCOW}’s coverage of multiple writing styles, registers, domains and genres (e.g., blogs, news articles, academic texts), ensures that our analyses reflect natural language in general.
Our combined datasets therefore contain close to 2M English sentences covering the whole set of 297 metaphorical vs. literal {\small VO}s.
Appendix \ref{tab:app:met-lit-count1} presents the full list of metaphorical and literal {\small VO}s and their sentence-level frequencies, reflecting how often each {\small VO} appears in context.

\paragraph{Features} Critically, we enriched our dataset with a range of cognitive and linguistic features: they serve as a basis for quantitative analyses of relationships between metaphorical vs. literal language and these properties. 
We extracted a total of 2,293 features using publicly-available {\small NLP} tools\footnote{\href{https://www.linguisticanalysistools.org/seance.html}{{\small SEANCE}},
\href{https://www.linguisticanalysistools.org/taales.html}{{\small TAALES}}, \href{https://www.linguisticanalysistools.org/taassc.html}{{\small TAASSC}}, \href{https://www.linguisticanalysistools.org/taaco.html}{{\small TAACO}}, \href{https://www.linguisticanalysistools.org/taaled.html}{{\small TAALED}}} which provide measures that are at the core of contrasting metaphorical vs. literal language usage, i.e., these features are related to lexical sophistication and diversity, text cohesion, syntactic complexity, and sentiment analysis (with integers in brackets referring to corresponding numbers of features): 
\vspace{+0.5mm}\\
{\small \textbf{SEANCE}} (1,271) assesses sentiment, emotion, and social cognition in texts using dictionaries and categories \citep{Crossley-et-al:2017}.\newline
{\small \textbf{TAALES}} (461) analyzes word frequency, range, concreteness, and psycholinguistic properties to assess lexical sophistication \citep{Kyle-Crossley:2015, Kyle-et-al:2018}.\newline
{\small \textbf{TAASSC}} (355) examines syntactic structures to measure complexity and variety in sentence construction \citep{Kyle:2016}.\newline
{\small \textbf{TAACO}} (168) measures text cohesion using lexical overlap, connectives, and semantic similarity \citep{Crossley-et-al:2016, Crossley-et-al:2019}.\newline
{\small \textbf{TAALED}} (38) computes lexical diversity metrics like type-token ratio, {\small MTLD} (Measure of Textual Lexical Diversity), and {\small HD-D} (Hypergeometric Distribution D) to evaluate vocabulary variation \citep{Kyle-et-al:2021}.


\section{Comparison of Metaphorical and Literal Verb-Objects: Approach}
\label{sec:approach}

For comparing metaphorical vs. literal {\small VO}s, this work addresses two research questions: 
\paragraph{RQ1} Are there consistent differences in \textbf{cognitive and linguistic feature patterns} between literal and metaphorical language (= \textbf{cross-pair})?
\paragraph{\textbf{RQ2}} Are there \textbf{specific {\small VO} pairs} showing a particularly strong contrast in cognitive and linguistic features between literal and metaphorical uses (=~\textbf{within-pair}), and do we find systematic patterns for literal- vs.\ metaphor-dominant {\small VO} pairs?

\paragraph{RQ1} To investigate how metaphorical and literal expressions differ in their contexts, we conducted pairwise comparisons of automatically extracted features across all sentences containing
our 297 {\small VO}s.
For each {\small VO}, we extracted contextual features using the {\small NLP} tools presented in Section \ref{sec:data}. To avoid the introduction of lexical bias, we masked both the verb and the object in all sentences by replacing them with {\small MASK} placeholders. 
All sentences containing the same {\small VO} were grouped and processed together as a single input to each tool. 
This allowed us to derive aggregated feature representations at the {\small VO} level, based solely on the surrounding context across all occurrences of the {\small VO}, but independent of the lexical identity of the verb and object themselves.
We then compared the feature scores between the metaphorical and literal variants of each pair, to investigate whether metaphorical sentences consistently differ from their literal counterparts along certain dimensions, e.g., whether they tend to exhibit higher emotional valence, lower cohesion, or greater syntactic complexity. 
We standardized all feature scores using z-score normalization to ensure comparability across scales. 

We conducted two levels of analysis. 
First, to assess (1) \textbf{cross-pair feature distinctions}, we computed the mean difference and conducted a Wilcoxon signed-rank test for each feature across all aligned metaphorical-literal {\small VO} pairs. 
Features were classified based on both statistical significance and effect size. Specifically, we considered a feature \textbf{important} if it showed a statistically significant difference $p < 0.05$ \textit{and} a substantial effect size defined as a mean difference of 
$\ge 1$ standard deviations following z-score normalization.
Features that were statistically significant but had smaller effect sizes were labeled as \textbf{significant but small}, indicating a consistent trend without a strong magnitude. Features with non-significant differences $p \geq 0.5$ were not considered further. 
We present the analyses and discussion in Sec. \ref{subsec:cross-pair-analysis}. 

\paragraph{RQ2} Second, we computed (2) \textbf{within-pair feature differences} to determine which {\small VO} pairs showed the largest divergence in their metaphorical vs.~literal usages. 
For each aligned {\small VO} pair, we calculated both signed and absolute differences across all normalized feature scores. We then computed the mean absolute difference across features per pair, using this value as an indicator of overall metaphoric--literal divergence. 
{\small VO} pairs with a mean absolute difference $\ge 1$ standard deviations were interpreted as \textbf{linguistically distinctive}; that is, their metaphorical and literal forms showed consistent contextual differences across multiple linguistic features. For these pairs, we also identified which features contributed most to the divergence, and whether the feature was higher in the literal or metaphorical variant. This analysis (Section \ref{subsec:within-pair-contrast}) allows us to explore the extent to which individual {\small VO} pairs show systematic variation in their surrounding contexts, as captured by the tools, thereby complementing our global feature-level comparisons.


\section{Analyses and Discussion}
\label{sec:analyses}

This section provides our analyses and discussions regarding our two research questions
These two main studies in Sections \ref{subsec:cross-pair-analysis} and \ref{subsec:within-pair-contrast} are preceded by a frequency analysis (Section \ref{subsec:freq}).

\subsection{Metaphorical vs. Literal Preferences}
\label{subsec:freq}

\begin{figure*}[ht!]
    \centering
    \includegraphics[width=\textwidth]{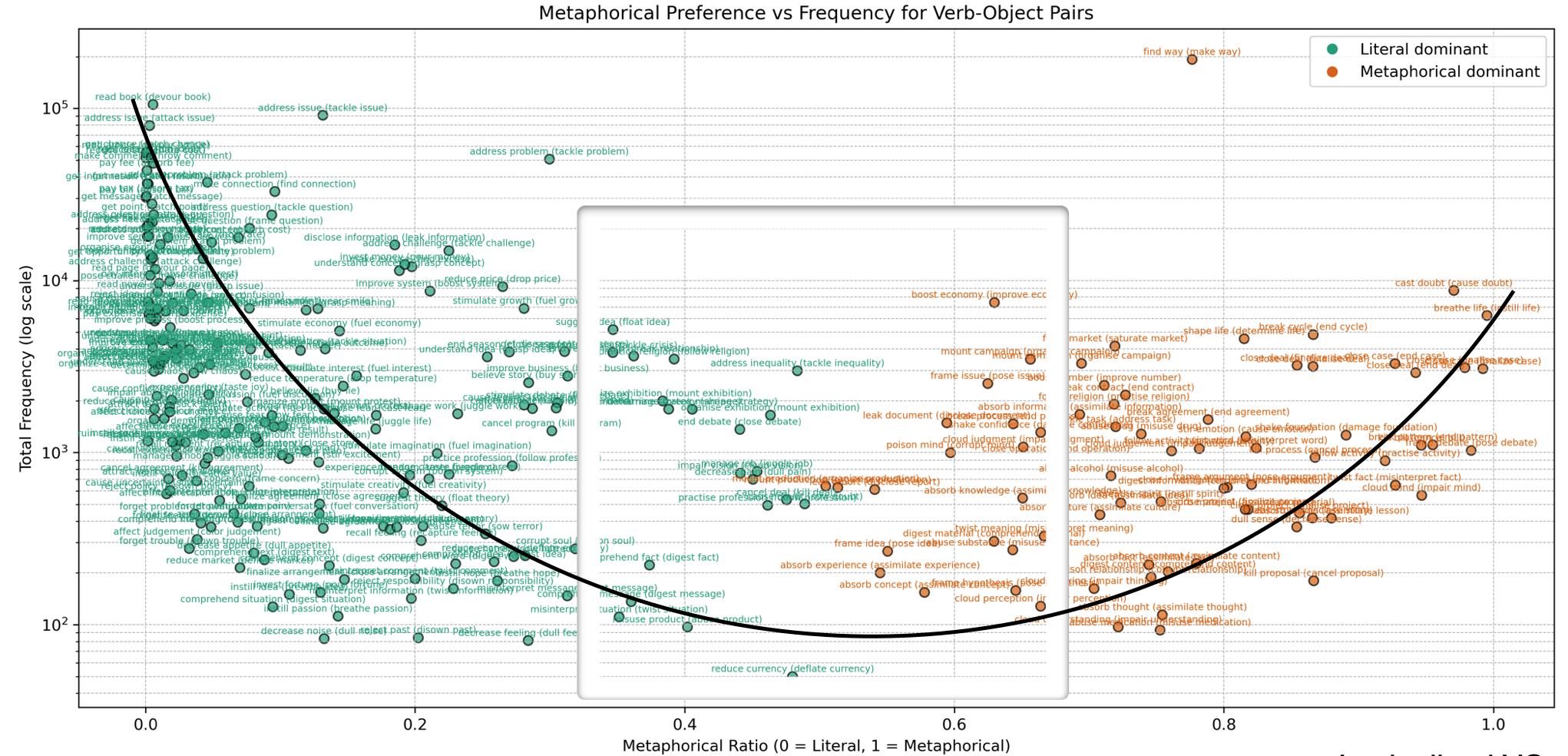}
    \caption{Metaphorical vs. literal {\small VO} preferences against total frequencies on a logarithmic scale, with points color-coded by dominance (green for literal-dominant, orange for metaphorical-dominant). The ratio ranges from 0 (entirely literal usage) to 1 (entirely metaphorical usage). Labels indicate the dominant {\small VO} followed by its corresponding counterpart in parenthesis, For example, met. {\small VO} \textit{breathe life} (lit. {\small VO} \textit{instill life}) is metaphorical dominant with a ratio of 1.0. For better readability, we report in Table \ref{tab:top10} the 10 most (i) frequency-balanced, (ii) literal-dominant and (iii) metaphor-dominant {\small VO} pairs.}
    \label{fig:vo-met-pref}
\end{figure*}

\begin{figure*}[h!]
    \centering
    \includegraphics[width=\textwidth]{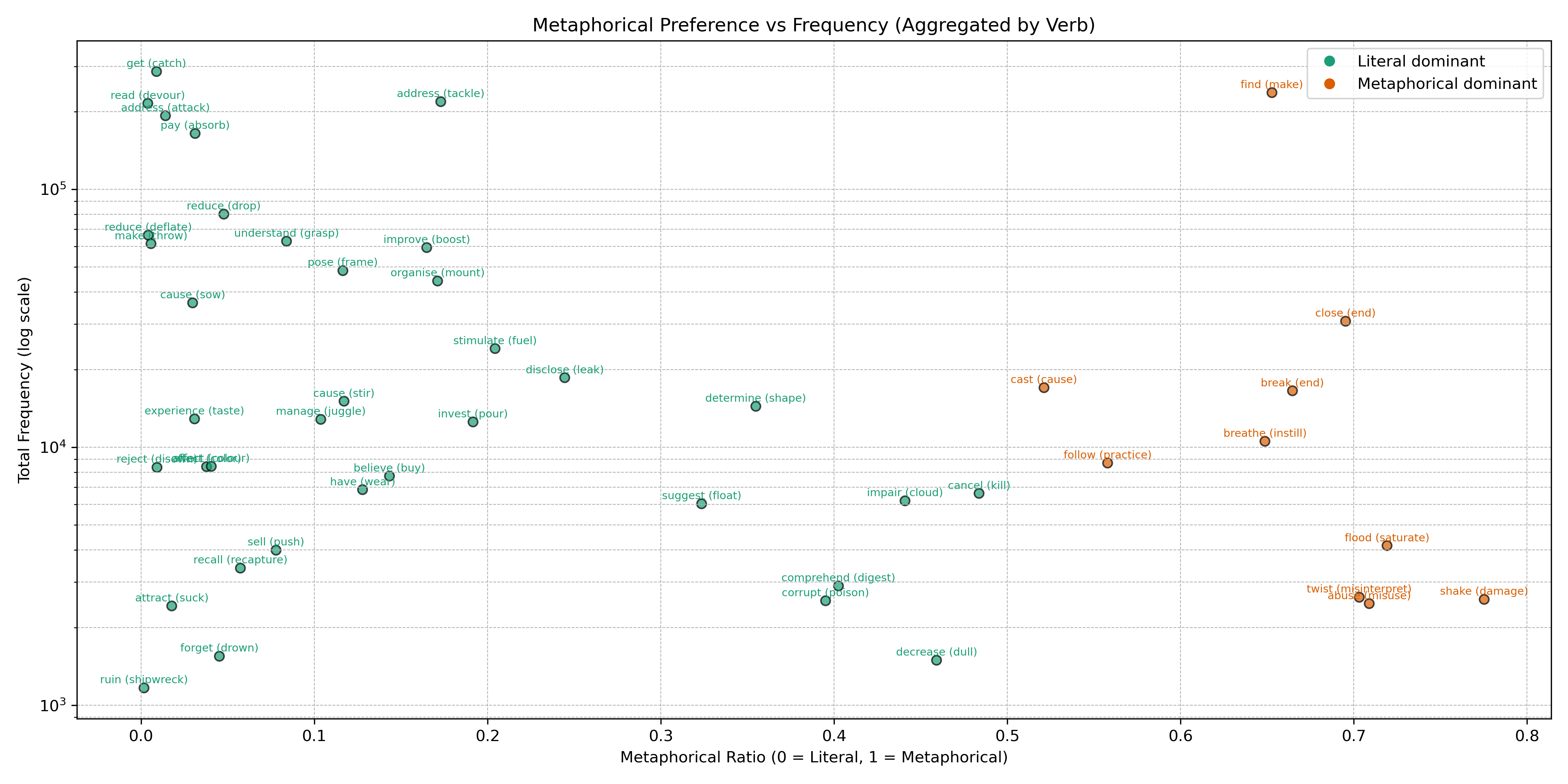}
    \caption{Metaphorical vs. literal verb-only preferences (independently of their direct objects) against total frequencies on a logarithmic scale, with points color-coded by dominance (green for literal-dominant, orange for metaphorical-dominant). The ratio ranges from 0 (entirely literal usage) to 1 (entirely metaphorical usage). Labels indicate the dominant verb followed by its corresponding counterpart in parenthesis.}
    \label{fig:vonly-met-pref}
\end{figure*}

\paragraph{Frequency} Table \ref{tab:vo_distribution} presents an overview of the distribution of literal and metaphorical sentences across our 297 {\small VO} pairs; in total, our dataset contains 2,058,787 sentences. In order to analyze metaphorical versus literal usage preferences, we defined a metaphorical ratio {\small $MET_{ratio}$} for each {\small VO} pair and also for each verb:
\vspace{+2mm}\\
%
%
\hspace*{+10mm}$MET_{ratio} = \frac{|MET|}{|MET| + |LIT|}$
\vspace{+3mm}\\
where {\small $|MET|$} and {\small $|LIT|$} denote the number of corpus occurrences of metaphorical and literal variants, respectively.
We included only pairs and verbs with a total frequency $>$ 10 to avoid instability from rare items.
Figures \ref{fig:vo-met-pref} and \ref{fig:vonly-met-pref} display scatter plots of metaphorical vs. literal preferences against total verb frequencies on a logarithmic scale.
These visualizations reveal both the overall distribution 
across our verbs and ({\small VO}) pairs along with their strong literal or metaphorical preferences, thus contextualizing the literal dominance observed in raw counts as a consequence of skewed frequency distributions (Table \ref{tab:vo_distribution}).


\begin{table}[h]
\centering
\begin{tabular}{lr}
\toprule
\textbf{Category} & \textbf{Count} \\
        \midrule
Total number of sentences & 2,058,787 \\
Sentences with literal {\small VO}s & 1,724,892 \\
Sentences with metaphorical {\small VO}s & 333,895 \\
Total number of {\small VO} pairs & 297 \\
{\small VO} pairs literal > metaphorical & 222 \\
{\small VO} pairs metaphorical > literal & 75 \\
{\small VO} pairs with equal frequency & 0 \\
\bottomrule
\end{tabular}
\caption{Distribution of literal and metaphorical sentences across  {\small VO} pairs in the dataset. The table reports the total number of sentences per category and the number of {\small VO} pairs for which one usage type predominates.}
\label{tab:vo_distribution}
\end{table}

Due to the visual density of Figure \ref{fig:vo-met-pref}, we report in Table \ref{tab:top10} the 10 most (i) frequency-balanced, (ii) literal-dominant and (iii) metaphor-dominant {\small VO} pairs.
Most targets (Figure \ref{fig:vo-met-pref}) are concentrated towards the left side ($ratio < 0.5$), indicating that literal usages (in green) tend to be more frequent overall for many {\small VO} pairs. As shown in Table \ref{tab:vo_distribution}, 222 of our {\small VO}s are literally dominant, such as \textit{read article} (vs. \textit{devour article}), \textit{address need} (vs. \textit{attack need}), \textit{pay bill} (vs. \textit{absorb bill}) (middle part of Table \ref{tab:top10}), compared to 75 {\small VO} pairs which are predominantly used metaphorically, i.e., cluster of orange points with a $ratio > 0.5$, such as \textit{cast doubt} (vs. \textit{cause doubt}), \textit{kill bill} (vs. \textit{cancel bill}), \textit{break pattern} (vs. \textit{end pattern}) (middle part of Table \ref{tab:top10}). The metaphorically-dominant {\small VO}s might represent expressions that are lexicalized in natural language (highly-conventionalized metaphors or \textit{dead} metaphors) and behave similarly to idiomatic expressions \citep{Lakoff-Johnson:1980, Bowdle-Gentner:2005}. 
Table \ref{tab:top10} further shows that literal-dominant {\small VO}s tend to exhibit larger absolute Log$_2$ values, indicating a more pronounced frequency imbalance than metaphor-dominant {\small VO}s, suggesting that literal usages often overwhelmingly outnumber their metaphorical counterparts, whereas metaphorical preferences are typically less extreme.

Points near the center ($0.35 < ratio < 0.65$) indicate pairs where metaphorical and literal uses occur with roughly equal frequencies, e.g., \textit{deflate} vs. \textit{reduce currency}, \textit{frame} vs. \textit{pose idea}, \textit{abuse} vs. \textit{misuse product} (top part of Table \ref{tab:top10}). These balanced cases are particularly interesting for our further qualitative analysis, as they may indicate more flexible, context-dependent interpretation (Section \ref{subsec:within-pair-contrast}).

\begin{table*}[t]
\centering
\begin{threeparttable}
\begin{tabular}{llrrrr}
\toprule
\textbf{Metaphorical} & \textbf{Literal} & \textbf{Met Count} & \textbf{Lit Count} & \textbf{Ratio} & \textbf{Log$_2$} \\
\midrule
\multicolumn{6}{l}{\textit{Top 10 frequency-balanced VO Pairs}} \\
\midrule
absorb concept & assimilate concept & 89 & 65 & 0.58 & 0.45 \\
absorb experience & assimilate experience & 109 & 91 & 0.55 & 0.26 \\
frame idea & pose idea & 147 & 120 & 0.55 & 0.29 \\
mount production & organise production & 322 & 305 & 0.51 & 0.07 \\
mount production & organize production & 322 & 316 & 0.51 & 0.03 \\
sow doubt & cause doubt & 246 & 257 & 0.49 & -0.06 \\
kill deal & cancel deal & 253 & 279 & 0.48 & -0.14 \\
deflate currency & reduce currency & 24 & 50 & 0.48 & -1.06 \\
abuse product & misuse product & 39 & 58 & 0.40 & -0.57 \\
twist situation & misinterpret situation & 39 & 72 & 0.35 & -0.88 \\

\midrule
\multicolumn{6}{l}{\textit{Top 10 VO pairs literal $<$ metaphor}} \\
\midrule
breathe life & instill life & 6,227 & 30 & 1.00 & 7.68 \\
close case & finalize case & 3,047 & 24 & 0.99 & 6.94 \\
frame debate & pose debate & 1,011 & 17 & 0.98 & 5.85 \\
close case & finalise case & 3,047 & 66 & 0.98 & 5.52 \\
cast doubt & cause doubt & 8,494 & 257 & 0.97 & 5.04 \\
break pattern & end pattern & 1,060 & 50 & 0.95 & 4.38 \\
cloud mind & impair mind & 534 & 30 & 0.95 & 4.11 \\
kill bill & cancel bill & 1,043 & 59 & 0.95 & 4.13 \\
close deal & end deal & 2,744 & 168 & 0.94 & 4.02 \\
twist fact & misinterpret fact & 599 & 47 & 0.93 & 3.66 \\

\midrule
\multicolumn{6}{l}{\textit{Top 10 VO pairs literal $>$ metaphor}} \\
\midrule
deflate cost & reduce cost & 12 & 51,650 & 0.00 & -12.07 \\
absorb bill & pay bill & 10 & 30,364  & 0.00 & -11.57 \\
devour article & read article &  22 & 55,121 & 0.00 & -11.29 \\
absorb tax & pay tax & 20 & 30,911 & 0.00 & -10.60 \\
attack need & address need &  20 & 20,487 & 0.00 & -10.00 \\
absorb fee & pay fee & 45 & 43,490 & 0.00 & -9.93 \\
catch chance & get chance & 67 & 55,978 & 0.00 & -9.71 \\
devour story & read story & 23 & 17,990 & 0.00 & -9.61 \\
catch result & get result & 47 & 36,573 & 0.00 & -9.60 \\
shipwreck career & ruin career & 2 & 1,165 & 0.00 & -9.19 \\
\bottomrule
\end{tabular}
\caption{Top {\small VO} pairs across distributional profiles: (i) the 10 most frequency-balanced pairs, (ii) the 10 pairs where metaphorical usage is more frequent than their literal counterparts, and (iii) the 10 pairs where literal usage is more frequent. \textbf{Met Count} and \textbf{Lit Count} indicate the number of sentences in which the metaphorical and literal {\small VO} occur, respectively. \textbf{Ratio} corresponds to the proportion of metaphorical usage (cf. Figure \ref{fig:vo-met-pref}). \textbf{Log$_2$} represents the base-2 logarithm of the ratio between metaphorical and literal frequencies, capturing the direction and magnitude of the imbalance.}
\label{tab:top10}
\end{threeparttable}
\end{table*}

Also, we observe a clear U-shaped pattern in the distribution of {\small VO} pairs in Figure \ref{fig:vo-met-pref}: pairs that are strongly literal or strongly metaphorical are substantially more frequent than pairs with a balanced literal–metaphorical ratio (see lit. \textit{read book} (upper left) and met. \textit{cast doubt} (upper right) vs. lit. \textit{practise profession} and met. \textit{absorb experience} (middle). This pattern reflects a usage-based effect: highly frequent verb–object constructions become lexicalized, stabilizing either toward predominantly literal meanings or toward conventional metaphorical meanings. In contrast, low-frequency pairs show less semantic stabilization and therefore exhibit more variable usage, clustering near the center of the ratio scale.

Finally, an important insight from the verb-only plot (Figure \ref{fig:vonly-met-pref}) is that examining verbs independently of their objects can be misleading. While the verb–object \textit{reduce currency} appears as a balanced phrase ($ ratio \ge 0.5$ and relatively low frequency) the verb \textit{reduce} on its own, aggregated across all objects, is strongly literal (ratio close to 0) with a very high frequency. This finding 
highlights how crucial it is to consider the verb and its object as a unit, as verbs behave very differently in terms of their frequency and metaphoricity when considered with or without their objects. 
\subsection{Cross-Pair Analysis: Feature-based Comparison of Metaphorical and Literal VOs}
\label{subsec:cross-pair-analysis}

In this section, we address our first research question:
\textbf{Are there consistent differences in \textbf{linguistic feature patterns} between literal and metaphorical language (= across {\small VO}s)}?

\paragraph{Affective Semantics ({\small SEANCE})} 
We observed 550 features that are statistically significant (with small effect). Features more prominent in literal uses (309) reflect positive affect, institutional structure, cognitive competence, and goal-directedness (valence, work and role-related lexicons, means–end procedural frames), as in \textit{"This ‌\textbf{bill} was ‌\textbf{paid} from an advance on a book whose unauthorised manuscript was published by another ally […]"} (lit. {\small VO}).
In contrast, metaphorical {\small VO} pairs were associated with 241 features related to negative affect, sensorimotor grounding, and social/personal vulnerability. These included emotion-laden categories (e.g., Fear, Disgust), embodied experience (e.g., Fall, Pain), and social identity markers (e.g., Race, RcEthic\_Lasswell -- a dictionary-based measure capturing the relative frequency of moral and ethical evaluation terms) as in \textit{"Fears, suspicions, resentments and hatred have \textbf{‌poisoned ‌relationships} across that divide in ways that threaten us all"} (met. {\small VO}). 
This pattern aligns with theories of metaphor as grounded in bodily and affective experience, often used to frame complex or disruptive states \citep{Lakoff-Johnson:1980, Gibbs:2006}.
The contrast supports a broad cognitive distinction: literal language tends to encode structured and abstract functionality (\textit{address question, reduce cost}, \textit{suggest idea, understand meaning}) and institutional roles (\textit{pay bill, organize conference}) \citep{Turner:1996, Boroditsky:2000}, while metaphor draws on vivid, sensorimotor grounding (\textit{digest idea, grasp risk)} and emotionally charged semantics (\textit{twist meaning, poison relationship)}, often serving to make abstract concepts experientially accessible \citep{Koevecses:2010, Mohammad-etal:2016}.

\paragraph{Lexical Sophistication ({\small TAALES})} Data containing literal {\small VO}s scored higher across a broad set of 147 frequency-, range-, and corpus-based features 
They show higher word and n-gram frequencies across major corpora ({\small SUBTLEX}us, {\small BNC}, {\small COCA}), broader contextual range, greater overlap with academic wordlists, suggesting that literal expressions occur in more common, general-purpose, and informative contexts, rather than in emotive ones.
We also observed higher scores on features for lexical decision times (e.g., {\small LD\_Mean\_RT\_CW} -- average lexical decision reaction time), indicating that literal language is easier to recognize and process.
In contrast, metaphorical expressions showed 73 features with higher scores on psycholinguistic dimensions such as imageability, concreteness, and meaningfulness ({\small MRC} norms), along with greater semantic richness (broader and more salient association strength, high {\small LSA}-based similarity). 
Notice, for example, the concrete, technical, and lexically familiar language used in the sentence \textit{"It’s the modifications to database permissions that \textbf{causes issues} with bbPress 2.2"} (lit. {\small VO}), whereas the sentence \textit{“Mikey’s stereotypes are holding him back and \textbf{coloring} his \textbf{judgement}}” (met. {\small VO}) contains more abstract, highly imageable language, plus another conventionalized metaphorical expression (\textit{smth. holding sb. back}). 
This contrast suggests that while literal language tends to rely on familiar, high-frequency lexical items optimized for ease of processing \citep{VanPetten-Kutas:1990, Dufau-et-al:2015}, metaphorical language engages more semantically rich, imageable, and conceptually salient terms \citep{Bowdle-Gentner:2005, Giora:2002}, potentially facilitating deeper conceptual associations and more nuanced interpretation. 

\paragraph{Syntactic Complexity ({\small TAASSC})} Quite a few features (33) are statistically significant and with a large effect, especially in contexts containing metaphorical {\small VO}s (26 for metaphorical vs.~7 for literal contexts). 
Contexts of literal {\small VO}s showed higher variability in syntactic roles such as passive subjects, agents, and indirect objects, as in \textit{"The care planning \textbf{system} in the home is currently being \textbf{improved}}" (lit. {\small VO}), suggesting flexible role-filling within relatively stable constructions. 
In contrast, metaphorical contexts scored higher on a larger set of features, including type-token ratios and lemma/construction frequency across multiple registers (academic, news, magazine, fiction), reflecting broader lexical and constructional diversity. 
The pattern holds with 171 significant but small-effect features, where literal contexts show more structural variability (e.g., in syntactic dependencies and construction frequency dispersion), reinforcing their flexibility in grammatical realization \citep{Fazly-etal:2009, Wierzba-et-al:2013}. 
Metaphorical contexts show a greater affinity for modifiers (e.g., possession, adverbs), and more conventionalized or lexically cohesive patterns (e.g., higher collexeme ratios, more frequent lemma-construction associations), suggesting semantic specificity and idiomatic usage  \citep{Gries-stefanowitsch:2004}.

\paragraph{Lexical Cohesion ({\small TAACO})} Metaphorical and literal language is also contrastive with regards to cohesion patterns. 
Among features showing significant and large effects, literal contexts score higher on measures such as lexical density, repetition of content and pronoun lemmas, suggesting stronger local cohesion and referential continuity. In contrast, metaphorical contexts show higher type-token ratios across a wide range of grammatical categories (e.g., nouns, verbs, adjectives, adverbs) and n-gram windows (bigram/trigram lemma {\small TTR}), indicating greater lexical diversity and a broader distribution of lexical items across the discourse.
For features with significant but small effects, contexts of literal {\small VO}s again show more consistent adjacent overlap across a variety of part-of-speech categories, as well as higher semantic similarity (e.g., {\small Word2Vec}, {\small LSA}) between adjacent segments. These suggest stronger sequential cohesion and thematic persistence. In contrast, metaphorical expressions scored higher on {\small MATTR} (Moving-Average Type-Token Ratio variant) and showed slightly more binary overlap for content and function words, as well as increased use of temporal and oppositional discourse markers and a higher pronoun-to-noun ratio, pointing to a more fragmented but stylistically varied cohesion profile.
This pattern aligns with previous findings from {\small TAASSC} and {\small TAALES}: literal language favors consistency, repetition, and higher cohesion at both the lexical and discourse levels, while metaphorical language tends to exhibit greater lexical and constructional variety \citep{Halliday-Hasan:1976}.

\paragraph{Lexical Diversity ({\small TAALED})} Similarly to the other tools, we observed significantly higher counts of tokens, types, and lexical density measures for literal contexts, suggesting a denser and more informationally rich lexical structure.
Also, literal contexts present moderate variation in word types (i.e., more varied vocabulary) within a relatively stable lexical range (i.e., within a coherent and semantically cohesive set of words).
In contrast, contexts of metaphorical {\small VO}s scored higher on alternative type-token ratio measures, indicating a greater variety of word types, especially for function words.
These findings, supported by smaller effects in measure like {\small MATTR}, {\small MSTTR} (Mean Segmental Type–Token Ratio), and {\small MTLD} (Measure of Textual Lexical Diversity), which provide length-controlled estimates of vocabulary variation, suggest that metaphorical language uses a more diverse and flexible vocabulary, particularly through varied use of function words that enforce cohesion and style \citep{Kimmel:2010, Piccirilli-SchulteImWalde:2022b}.

\begin{figure*}[ht]
    \centering
    \begin{subfigure}[b]{0.49\textwidth}
        \centering
        \includegraphics[width=\textwidth]{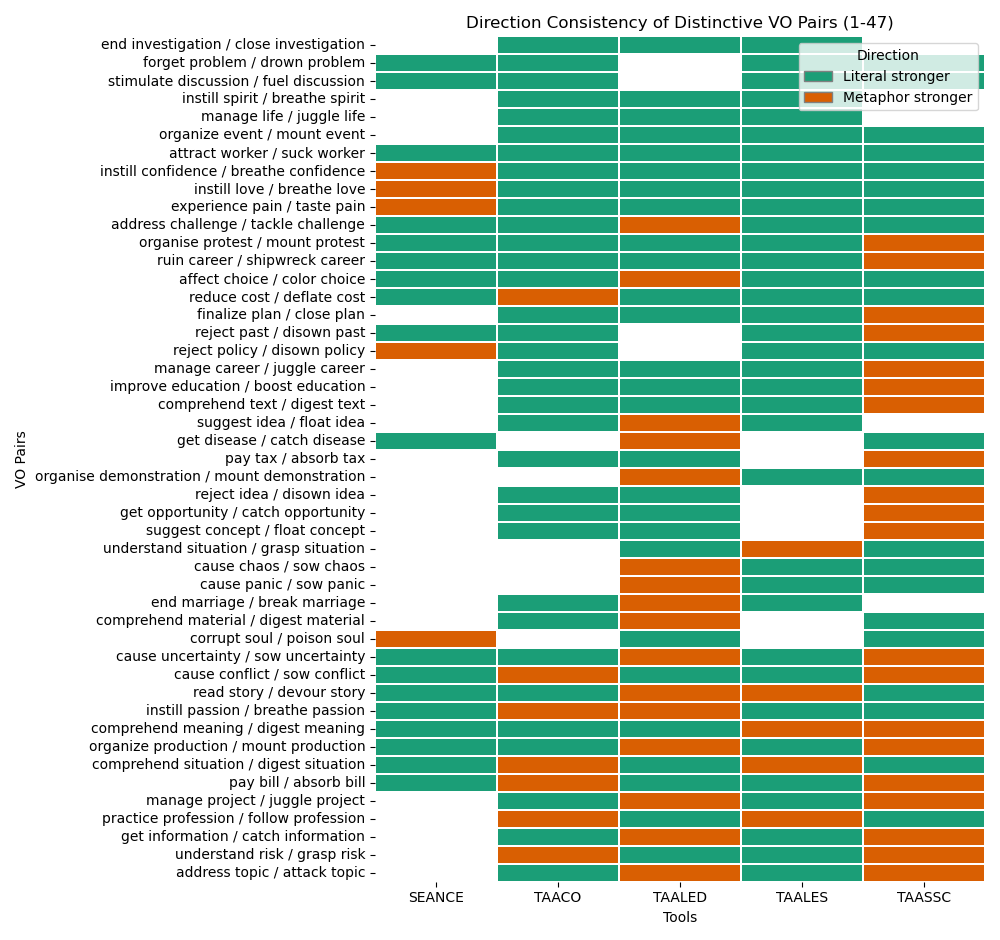} 
    \end{subfigure}
    \hfill
    \begin{subfigure}[b]{0.49\textwidth}
        \centering
        \includegraphics[width=\textwidth]{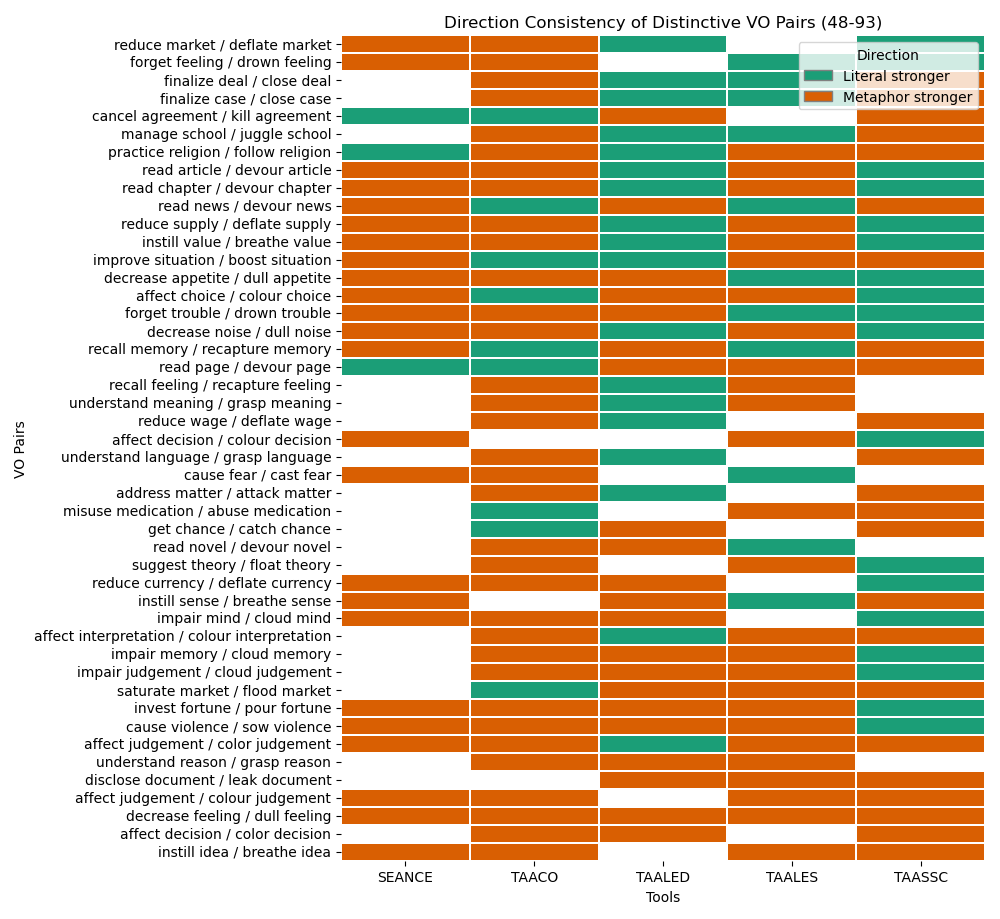} 
    \end{subfigure}
    \caption{Tool-level aggregation of distributional dominance  for 93 clearly-distinctive {\small VO} pairs (mean absolute difference $\ge 1$ standard deviations in at least three tools, detailed methodology explained in Sections \ref{sec:approach} and \ref{subsec:within-pair-contrast}). Feature-level contrasts were collapsed into tool-level directional judgments (literal-dominant in green vs. metaphor-dominant in orange) for {\small VO} pairs meeting the distinctiveness criterion in at least three tools. {\small VO} pairs are sorted top to bottom from mostly literal-dominant to mostly metaphor-dominant. For example, the {\small VO} pair \textit{instill idea vs. breathe idea} behaves significantly more "metaphorical-like" across features of four tools ({\small SEANCE, TAACO, TAALES} and {\small TAASSC}), i.e., the metaphorical variant \textit{breathe idea} introduces robust distributional differences detectable across linguistic features (structurally, lexically, discourse-wise, etc.) Sections \ref{sec:approach} and \ref{subsec:cross-pair-analysis} describe the linguistic aspects captures by the tools.} 
    \label{fig:direction-consistency}
\end{figure*}

\subsection{Within-Pair Analysis: Feature Differences between Metaphorical and Literal VOs}
\label{subsec:within-pair-contrast}

In Section \ref{subsec:cross-pair-analysis}, we took a \textbf{cross-pair} approach and looked at the overall contrast between metaphorical vs. literal {\small VO}s (= all metaphorical {\small VO}s vs. all literal {\small VO}s). In contrast, we now zoom into the contrast \textbf{within-pair}, that is to say we focus on each individual {\small VO} pair (e.g., \textit{float idea} vs. \textit{suggest idea}) and all the respective sentences the {\small VO}s appear in.
Our goal is to determine whether our near-synonymous literal and metaphorical {\small VO}s show substantial internal 
differences, and whether such differences are systematic across pairs or rather pair-specific.
We are therefore addressing our second research question: \textbf{Among the most distinctive {\small VO} pairs, which ones -- if any -- stand out for showing especially strong contrasts in linguistic features between their literal and metaphorical uses?}

\paragraph{Feature-level differences} For each {\small VO} pair and each feature, we computed the \textbf{signed standardized difference}:
\vspace{+2mm}\\
\hspace*{+10mm}$ signed\_diff = z_{metaphor} - z_{literal}$
\vspace{+3mm}\\
where z-scores were obtained using a single StandardScaler fitted on the combined literal and metaphorical data to ensure a shared reference distribution. Positive values indicate higher feature values for the metaphorical variant, while negative values indicate higher values for the literal variant.
We also computed the \textbf{absolute difference} $|signed\_diff|$ capturing magnitude differences irrespective of direction.

\paragraph{Aggregation across features:}
For each {\small VO} pair, we then computed the \textbf{mean absolute difference} across all features to quantify the overall distinctiveness of each pair, as well as the \textbf{mean signed difference} across features, capturing the overall directional bias (literal-dominant vs. metaphor-dominant).
{\small VO} pairs were then classified as \textbf{clearly distinctive} if their mean absolute difference equalled or exceeded a threshold of 1 standard deviations. This threshold identifies pairs whose literal and metaphorical realizations differ substantially across all features.
For each pair, we report: the (i) mean absolute difference, (ii) features (across tools) exceeding the defined threshold, and (iii) direction of dominance (literal vs. metaphorical)
This step does not assume that all features move in the same direction; rather, it quantifies the overall degree of distributional divergence between the two variants.
Finally, we can compute the directional classification. 
We define as \textbf{literal-dominant} the pairs with a negative mean signed difference and \textbf{metaphor-dominant} the ones with a positive mean signed difference.
Again, this classification reflects aggregate tendencies across all features. A metaphor-dominant pair does not imply that every feature is higher in the metaphorical variant, but that the balance of standardized feature differences favors that variant. 


Given the high dimensionality of the overall feature space ($> 2,000$ features across the five {\small NLP} tools we use), visualizing feature-level contrasts per distinctive {\small VO} pair separately for each tool proved to be difficult to interpret.
As a result, to improve readability and interpretability, while assessing robustness of our approach,
we identified {\small VO} pairs that met our two distinctiveness criterion: (1) a mean absolute difference $\ge 1$ standard deviations (2) in at least three of the five tools. 
For each tool, {\small VO} pairs were further classified as literal-dominant or metaphor-dominant based on the mean signed standardized difference across features. 

We present in Figure \ref{fig:direction-consistency} the aggregations of feature-level contrasts into tool-level directional judgments (literal-dominant in green vs. metaphor-dominant in orange), allowing us to assess whether metaphor–literal asymmetries persist across our independent linguistic feature set. 
We observe a lot of heterogeneity: according to our strict criteria, 93 {\small VO} pairs proved to be clearly distinctive, with only a subset of {\small VO} pairs actually showing consistent dominance patterns across tools.
(i) Seven {\small VO}s are literal-dominant across tools (green rows), such as \textit{forget} vs. \textit{drown problem}, \textit{organize} vs. \textit{mount event}, \textit{attract} vs. \textit{suck worker}.
(ii) Six {\small VO}s are mostly metaphor-dominant (orange rows), such as \textit{instill} vs. \textit{breathe idea}, \textit{understand} vs. \textit{grasp meaning}, \textit{disclose} vs. \textit{leak document}.
(iii) The rest of them -- 80 {\small VO}s -- actually show mixed patterns, where no consistent directional bias emerges. These mixed {\small VO}s are neither literal nor metaphor dominant, e.g., \textit{manage} vs. \textit{juggle project}, \textit{address} vs. \textit{attack topic}, meaning that different linguistic dimensions pull in different directions. For example, consider the {\small VO} pair \textit{pay bill} vs. \textit{absorb bill}: the metaphorical variant \textit{absorb bill} is on average dominant across {\small TAACO} and {\small TAASSC}, for which features related to lexical cohesion and syntactic complexity, respectively, scored higher than for its literal counterpart \textit{pay bill}. On the contrary, features measuring affective semantics ({\small SEANCE}), lexical sophistication ({\small TAALES}) and lexical diversity ({\small TAALED}) are pulling towards the literal variant \textit{pay bill}.

Our analysis suggests that metaphor–literal differences do not follow stable, recurring linguistic patterns. Instead, distributional asymmetries are pair-specific, with each {\small VO} pair showing its own pattern of divergence.
Moreover, directional patterns are not consistent even across contrasts involving the same literal–metaphorical verb pair. For instance, the contrast between \textit{instill} and \textit{breathe} is literal-dominant in combination with the direct object \textit{spirit}, but mostly metaphor-dominant with \textit{idea}. 
This finding indicates that distributional asymmetries are once again (as mentioned in Section \ref{subsec:freq}) not verb-driven but construction-specific: the unit of divergence is the verb–object pairing rather than the lexical verb itself. 
\section{Conclusion}

This study provides a large-scale, feature-based comparison of metaphorical and literal verb–object constructions in natural English sentences. Leveraging nearly two million corpus instances and over 2,200 linguistic features derived from five complementary {\small NLP} tools, we investigated both cross-pair and within-pair patterns of divergence.

At the \textbf{cross-pair} level, metaphorical and literal language show consistent distributional tendencies. Literal contexts are characterized by higher lexical frequency, stronger local cohesion, and more structurally regular patterns, suggesting stability, informativeness, and easier processing. 
In contrast, metaphorical contexts display greater affective intensity, sensorimotor grounding, lexical diversity, and constructional specificity, aligning with theoretical accounts that view metaphor as experientially grounded and semantically enriching.

In contrast, our \textbf{within-pair} analyses reveal that the cross-pair tendencies do not translate into uniform pair-level behavior. Although a subset of {\small VO} pairs shows strong and robust divergence (literal or metaphorical dominant), most pairs present mixed dominance patterns across linguistic dimensions. Moreover, contrasts are construction-specific rather than verb-driven: the same verb may pattern differently depending on its object, indicating that metaphoricity emerges at the level of the verb–object construction pairing rather than the lexical verb alone.

Our findings have both theoretical and methodological implications. Theoretically, they challenge the assumption of stable linguistic signatures of metaphoricity and instead support a usage-based, construction-sensitive view of figurative language. Methodologically, our results demonstrate the value of aggregating multidimensional feature profiles while preserving pair-level granularity. Future work may extend this approach to cross-linguistic settings, translation studies, and predictive modeling, further integrating linguistic theory with large-scale computational analysis. As a matter of fact, our follow-up study makes use of the extracted linguistic features and the knowledge learned from this work as input to machine learning models in order to observe the important features for predicting metaphorical vs. literal language.


\section{Limitations}

Several limitations should be acknowledged. 
First, our analysis is restricted to the English language, and the patterns observed in this work may not generalize to other languages with different morphological, syntactic, or metaphorical conventions, especially given the fact that the {\small NLP} tools we used are built on and for the English language. 
Second, although the corpus covers multiple genres (e.g., fiction, academic, web), it represents a single source, and corpus-specific distributional biases cannot be ruled out. 
Third, our dataset comprises 297 verb–object pairs, which, even though systematically constructed, cannot be considered fully representative of metaphorical or literal language as a whole. 
Finally, the very large number of extracted features ($>2,200$) made it necessary to aggregate and enforce thresholding procedures to ensure readability and interpretability: while this multidimensional approach enables broad coverage, it may also obscure more fine-grained effects at the level of individual features or verb-object features.

\section{Acknowledgments}
This research was supported by the DFG Research Grants SCHU 2580/4-1 (\textit{MUDCAT: Multimodal Dimensions and Computational Applications of Abstractness}) and SCHU 2580/7-1 and FR 2829/8-1 (\textit{MeTRapher: Learning to Translate Metaphors}). Prisca Piccirilli is also supported by the Studienstiftung des deutschen Volkes.
We are grateful to Annerose Eichel, Neele Falk and the {\small IMS} SemRel research group for helpful discussions, suggestions and feedback regarding versions of this work. We thank Sven Naber for his assistance with data extraction and processing. We would also like to thank the anonymous reviewers for their constructive feedback.

\section{Bibliographical References}
\label{sec:reference}

\bibliographystyle{lrec2026-natbib}
\bibliography{PriscaPicci}

@article{Boroditsky:2000,
  author  = {Lera Boroditsky},
  title   = {{Metaphoric Structuring: Understanding Time Through Spatial Metaphors}},
  journal = {Cognition},
  year    = {2000},
  doi = {https://doi.org/10.1016/S0010-0277(99)00073-6},
  volume  = {75},
  number  = {1},
  pages   = {1--28}
}

@article{Bowdle-Gentner:2005,
author = {Bowdle, Brian and Gentner, Dedre},
year = {2005},
month = {01},
pages = {193-216},
title = {{The Career of Metaphor}},
volume = {112},
journal = {Psychological Review},
doi = {10.1037/0033-295X.112.1.193}
}

@article{Cameron-et-al:2003,
author = {Cameron, Lynne and Deignan, Alice},
year = {2003},
month = {07},
pages = {},
title = {{Combining Large and Small Corpora to Investigate Tuning Devices Around Metaphor in Spoken Discourse}},
volume = {18},
journal = {Metaphor and Symbol},
doi = {10.1207/S15327868MS1803_02}
}

@article{Crossley-et-al:2016,
  author    = {Scott A. Crossley and Kristopher Kyle and Danielle S. McNamara},
  title     = {{The Tool for the Automatic Analysis of Text Cohesion (TAACO): Automatic Assessment of Local, Global, and Text Cohesion}},
  journal   = {Behavior Research Methods},
  volume    = {48},
  number    = {4},
  pages     = {1227--1237},
  year      = {2016},
  doi       = {10.3758/s13428-015-0651-7}
}

@article{Crossley-et-al:2017,
  author    = {Scott A. Crossley and Kristopher Kyle and Danielle S. McNamara},
  title     = {{Sentiment Analysis and Social Cognition Engine (SEANCE): An Automatic Tool for Sentiment, Social Cognition, and Social Order Analysis}},
  journal   = {Behavior Research Methods},
  volume    = {49},
  number    = {3},
  pages     = {803--821},
  year      = {2017},
  doi       = {10.3758/s13428-016-0743-z}
}

@article{Crossley-et-al:2019,
  author    = {Scott A. Crossley and Kristopher Kyle and Mihai Dascalu},
  title     = {{The Tool for the Automatic Analysis of Cohesion 2.0: Integrating Semantic Similarity and Text Overlap}},
  journal   = {Behavior Research Methods},
  volume    = {51},
  number    = {1},
  pages     = {14--27},
  year      = {2019},
  doi       = {10.3758/s13428-018-1142-4}
}

@article{Dufau-et-al:2015,
author = {Dufau, Stephane and Grainger, Jonathan and Midgley, Katherine and Holcomb, Phillip},
year = {2015},
month = {11},
pages = {},
title = {{A Thousand Words Are Worth a Picture: Snapshots of Printed-Word Processing in an Event-Related Potential Megastudy}},
volume = {26},
journal = {Psychological Science},
doi = {10.1177/0956797615603934}
}

@article{Fazly-etal:2009,
    title = {{Unsupervised Type and Token Identification of Idiomatic Expressions}},
    author = "Fazly, Afsaneh  and
      Cook, Paul  and
      Stevenson, Suzanne",
    journal = "Computational Linguistics",
    volume = "35",
    number = "1",
    year = "2009",
    pages = "61--103",
    url = "https://aclanthology.org/J09-1005",
}

@Article{Gibbs:1989,
  author = 	 {Raymond W. Gibbs},
  title = 	 {Understanding and {L}iteral {M}eaning},
  journal = 	 {Cognitive Science},
  year = 	 {1989},
  volume = 	 {13},
  number =       {2},
  pages = 	 {243--251},
  url="https://onlinelibrary.wiley.com/doi/pdf/10.1207/s15516709cog1302_5",
}

@article{Gibbs:2006,
	author = {Raymond W. Gibbs},
	journal = {Mind \& Language},
	pages = {434-458},
	title = {{Metaphor Interpretation as Embodied Simulation}},
	number = {3},
	volume = {21},
	year = {2006},
	doi = {10.1111/j.1468-0017.2006.00285.x},
}

@article{Giora:2002,
author = {Giora, Rachel},
year = {2002},
month = {04},
pages = {487-506},
title = {{Literal vs. Figurative Language: Different or Equal}},
volume = {34},
journal = {Journal of Pragmatics},
doi = {10.1016/S0378-2166(01)00045-5}
}

@article{Gries-stefanowitsch:2004,
author = {Gries, Stefan and Stefanowitsch, Anatol},
year = {2004},
month = {02},
pages = {97-129},
title = {{Extending Collostructional Analysis: A Corpus-Based Approach to Alternations}},
volume = {9},
journal = {International Journal of Corpus Linguistics},
doi = {10.1075/ijcl.9.1.06gri}
}

@Book{Halliday-Hasan:1976,
  author = 	 {Michael A.K. Halliday and Ruqaiya Hasan},
  title = 	 {{Cohesion in English}},
  publisher = 	 {Longman},
  year = 	 {1976},
  address = 	 {},
}

@article{Kimmel:2010,
author = {Kimmel, Michael},
year = {2010},
month = {01},
pages = {97-115},
title = {{Why We Mix Metaphors (and Mix Them Well): Discourse Coherence, Conceptual Metaphor, and Beyond}},
volume = {42},
journal = {Journal of Pragmatics},
doi = {10.1016/j.pragma.2009.05.017}
}

@Book{Koevecses:2010,
  author = 	 {Zoltan K\"ovecses},
  title = 	 {{Metaphor: A Practical Introduction}},
  publisher = 	 {Oxford University Press},
  year = 	 {2010},
  edition =      {2nd},
  address = 	 {New York},
}

@article{Kyle-Crossley:2015,
  author    = {Kristopher Kyle and Scott A. Crossley},
  title     = {{Automatically Assessing Lexical Sophistication: Indices, Tools, Findings, and Application}},
  journal   = {TESOL Quarterly},
  volume    = {49},
  number    = {4},
  pages     = {757--786},
  year      = {2015},
  doi       = {10.1002/tesq.194}
}

@phdthesis{Kyle:2016,
  author       = {Kristopher Kyle},
  title        = {{Measuring Syntactic Development in L2 Writing: Fine-Grained Indices of Syntactic Complexity and Usage-Based Indices of Syntactic Sophistication}},
  school       = {Georgia State University},
  year         = {2016},
  type         = {Ph.D. Dissertation},
  url          = {https://scholarworks.gsu.edu/alesl_diss/33/}
}

@article{Kyle-et-al:2018,
  author    = {Kristopher Kyle and Scott A. Crossley and Charles Berger},
  title     = {{The Tool for the Analysis of Lexical Sophistication (TAALES): Version 2.0}},
  journal   = {Behavior Research Methods},
  volume    = {50},
  number    = {3},
  pages     = {1030--1046},
  year      = {2018},
  doi       = {10.3758/s13428-017-0924-4}
}

@article{Kyle-et-al:2021,
  author    = {Kristopher Kyle and Scott A. Crossley and Scott Jarvis},
  title     = {{Assessing the Validity of Lexical Diversity Using Direct Judgements}},
  journal   = {Language Assessment Quarterly},
  volume    = {18},
  number    = {2},
  pages     = {154--170},
  year      = {2021},
  doi       = {10.1080/15434303.2020.1844205}
}

@Book{Lakoff-Johnson:1980,
  author = 	 {Lakoff, George and Johnson, Mark},
  title = 	 {Metaphors We Live By},
  publisher = 	 {University of Chicago Press},
  address = {Chicago},
  year = 	 {1980},
}

@inproceedings{Mohammad-etal:2016,
    title = {{Metaphor as a Medium for Emotion: An Empirical Study}},
    author = "Mohammad, Saif  and
      Shutova, Ekaterina  and
      Turney, Peter",
    booktitle = "Proceedings of the Fifth Joint Conference on Lexical and Computational Semantics",
    month = aug,
    year = "2016",
    address = "Berlin, Germany",
    publisher = "Association for Computational Linguistics",
    pages = "23--33",
    url = "https://aclanthology.org/S16-2003",
}

@article{Ortony:1975,
author = "Ortony, Andrew",
title = {{Why Metaphors Are Necessary and Not Just Nice}},
year = "1975",
language = "English (US)",
volume = "25",
pages = "45--53",
journal = "Educational Theory",
publisher = "Wiley-Blackwell",
number = "1",
url="https://users.cs.northwestern.edu/~ortony/Andrew_Ortony_files/1975\%20-\%20Why\%20metaphors\%20necessary.pdf",
}

@inproceedings{Piccirilli-SchulteImWalde:2021,
  author    = {Piccirilli, Prisca and {Schulte im Walde}, Sabine},
  title     = {{Synonymous Pairs of Metaphorical and Literal Expressions in Context: An Empirical Study and Dataset \textit{to tackle} or \textit{to address the question}}},
  booktitle = {Proceedings of the Workshop DiscAnn},
  year      = {2021},
  address   = {T\"ubingen, Germany},
  url="https://www.ims.uni-stuttgart.de/documents/team/schulte/publications/workshop/discann-21.pdf",
}

@inproceedings{Piccirilli-SchulteImWalde:2022b,
    title = {{What Drives the Use of Metaphorical Language? Negative Insights from Abstractness, Affect, Discourse Coherence and Contextualized Word Representations}},
    author = "Piccirilli, Prisca  and
      Schulte Im Walde, Sabine",
    editor = "Nastase, Vivi  and
      Pavlick, Ellie  and
      Pilehvar, Mohammad Taher  and
      Camacho-Collados, Jose  and
      Raganato, Alessandro",
    booktitle = "Proceedings of the 11th Joint Conference on Lexical and Computational Semantics",
    month = jul,
    year = "2022",
    address = "Seattle, Washington",
    publisher = "Association for Computational Linguistics",
    url = "https://aclanthology.org/2022.starsem-1.26/",
    doi = "10.18653/v1/2022.starsem-1.26",
    pages = "299--310"
}

@inproceedings{Piccirilli-et-al:2024,
    title = "{{VOLIMET}: A Parallel Corpus of Literal and Metaphorical Verb-Object Pairs for {E}nglish{--}{G}erman and {E}nglish{--}{F}rench}",
    author = "Piccirilli, Prisca  and
      Fraser, Alexander  and
      Schulte im Walde, Sabine",
    editor = "Bollegala, Danushka  and
      Shwartz, Vered",
    booktitle = "Proceedings of the 13th Joint Conference on Lexical and Computational Semantics (*SEM 2024)",
    month = jun,
    year = "2024",
    address = "Mexico City, Mexico",
    publisher = "Association for Computational Linguistics",
    url = "https://aclanthology.org/2024.starsem-1.18/",
    pages = "222--237",
}

@inproceedings{Schafer:2015,
  author    = {Roland Sch{\"a}fer},
  title     = {{Processing and Querying Large Web Corpora with the COW14 Architecture}},
  booktitle = {Proceedings of the 3rd Workshop on Challenges in the Management of Large Corpora},
  editor    = {Piotr Bański and Hanno Biber and Evelyn Breiteneder and Marc Kupietz and Harald L{\"u}ngen and Andreas Witt},
  publisher = {Institut f{\"u}r Deutsche Sprache},
  address   = {Mannheim, Germany},
  url       = {https://nbn-resolving.org/urn:nbn:de:bsz:mh39-38367},
  pages     = {28--34},
  year      = {2015},
}

@inproceedings{Schafer-Bildhauer:2012,
    title = {{Building Large Corpora from the Web Using a New Efficient Tool Chain}},
    author = {Sch{\"a}fer, Roland  and
      Bildhauer, Felix},
    booktitle = "Proceedings of the 8th International Conference on Language Resources and Evaluation",
    year = "2012",
    address = "Istanbul, Turkey",
    publisher = "European Language Resources Association",
    url = "http://www.lrec-conf.org/proceedings/lrec2012/pdf/834_Paper.pdf",
    pages = "486--493",
}

@Article{Schaeffner:2004,
  author =       {Sch\"affner, Christina},
  title = 	 {{Metaphor and Translation: Some Implications of a Cognitive Approach}},
  journal = 	 {Journal of Pragmatics},
  year = 	 {2004},
  volume = 	 {36(7)},
  pages = 	 {1253--1269},
  url="https://reader.elsevier.com/reader/sd/pii/S0378216604000244?token=A5BCBEBAC3CF0190D1CE5C134686B62168884DDF18D41D57D343D5A978065F5404116B0B307408D015C2CC24F3CF0F25&originRegion=eu-west-1&originCreation=20220511163221",
}

@inproceedings{Shutova:2010,
    title = {{Automatic Metaphor Interpretation as a Paraphrasing Task}},
    author = "Shutova, Ekaterina",
    booktitle = "Proceedings of the Annual Conference of the North {A}merican Chapter of the Association for Computational Linguistics: Human Language Technologies: ",
    month = jun,
    year = "2010",
    address = "Los Angeles, California",
    publisher = "Association for Computational Linguistics",
    pages = "1029--1037",
    url = "https://aclanthology.org/N10-1147",
}

@Article{Steen:2004,
  author = 	 {Gerard J. Steen},
  title = 	 {{Can Discourse Properties of Metaphor Affect Metaphor Recognition?}},
  journal = 	 {Journal of Pragmatics},
  year = 	 {2004},
  volume = 	 {36(7)},
  pages = 	 {1295--1313},
  doi="10.1016/J.PRAGMA.2003.10.014",
}

@inproceedings{Stowe-et-al:2022,
    title = {{{IMPLI}: Investigating {NLI} Models{'} Performance on Figurative Language}},
    author = "Stowe, Kevin  and
      Utama, Prasetya  and
      Gurevych, Iryna",
    booktitle = "Proceedings of the 60th Annual Meeting of the Association for Computational Linguistics",
    month = may,
    year = "2022",
    address = "Dublin, Ireland",
    publisher = "Association for Computational Linguistics",
    url = "https://aclanthology.org/2022.acl-long.369",
    pages = "5375--5388",
}

@book{Turner:1996,
  author    = {Mark Turner},
  title     = {{The Literary Mind: The Origins of Thought and Language}},
  year      = {1996},
  publisher = {Oxford University Press},
  address   = {Oxford}
}

@Article{vandenBroek:1981,
  author = 	 {{van den Broek}, Raymond},
  title = 	 {{The Limits of Translatability Exemplified by Metaphor Translation}},
  journal = 	 {Poetics Today},
  publisher =    {Duke University Press},
  year = 	 {1981},
  volume = 	 {1},
  number = 	 {4},
  pages = 	 {73--87},
  doi={10.2307/1772487}
}

@article{VanPetten-Kutas:1990,
author = {Van Petten, Cyma and Kutas, Marta},
year = {1990},
month = {08},
pages = {380-93},
title = {{Interactions Between Sentence Context and Word Frequency in Event-related Potentials}},
volume = {18},
journal = {Memory \& cognition},
doi = {10.3758/BF03197127}
}

@article{Wierzba-et-al:2013,
author = {Wierzba, Marta and Brown, J. and Fanselow, Gisbert},
year = {2023},
month = {04},
pages = {1-38},
title = {{The Syntactic Flexibility of German and English Idioms: Evidence from Acceptability Rating Experiments}},
volume = {60},
journal = {Journal of Linguistics},
doi = {10.1017/S0022226723000105},
}

@Article{Blasko:99,
  author = 	 {Dawn G. Blasko},
  title = 	 {{"Only the Tip of the Iceberg": {W}ho Understands what about Metaphor?}},
  journal = 	 {Journal of Pragmatics},
  year = 	 {1999},
  doi = {https://doi.org/10.1016/S0378-2166(99)00009-0},
  volume = 	 {31},
  pages = 	 {1675--1683},
}

@Article{Glucksberg:03,
  author = 	 {Sam Glucksberg},
  title = 	 {{The Psycholinguistics of Metaphor}},
  journal = 	 {TRENDS in Cognitive Science},
  year = 	 {2003},
  doi = {10.1016/S1364-6613(02)00040-2},
  volume = 	 {7},
  number = 	 {2},
  pages = 	 {92--96},
}


\appendix

\section{Supplementary Materials}
\label{sec:appendix}

\subsection{Met vs. Lit Verb-Object Pairs}

\onecolumn
\begin{longtable}{llrr}
\caption{Metaphorical vs. Literal Verb--Object Pairs with Corpus Frequencies (N=297)}\\
\toprule
\textbf{Metaphorical {\small VO}} & \textbf{Literal {\small VO}} & \textbf{Met Count} & \textbf{Lit Count}\\
\midrule
\endfirsthead

\toprule
\textbf{Metaphorical {\small VO}} & \textbf{Literal {\small VO}} & \textbf{Met Count} & \textbf{Lit Count}\\
\midrule
\endhead

\bottomrule
\endfoot

absorb knowledge & assimilate knowledge & 354 & 190 \\
absorb information & assimilate information & 1,154 & 511 \\
absorb idea & assimilate idea & 369 & 141 \\
absorb culture & assimilate culture & 308 & 127 \\
absorb material & assimilate material & 382 & 85 \\
absorb lesson & assimilate lesson & 366 & 50 \\
absorb fact & assimilate fact & 166 & 57 \\
absorb content & assimilate content & 177 & 50 \\
absorb experience & assimilate experience & 109 & 91 \\
absorb concept & assimilate concept & 89 & 65 \\
absorb thought & assimilate thought & 86 & 28 \\
absorb cost & pay cost & 1,226 & 16,540 \\
absorb fee & pay fee & 45 & 43,490 \\
absorb bill & pay bill & 10 & 30,364 \\
absorb tax & pay tax & 20 & 30,911 \\
absorb debt & pay debt & 61 & 21,344 \\
absorb interest & pay interest & 178 & 9,713 \\
absorb expense & pay expense & 40 & 5,774 \\
abuse alcohol & misuse alcohol & 525 & 208 \\
abuse drug & misuse drug & 947 & 335 \\
abuse substance & misuse substance & 175 & 97 \\
abuse medication & misuse medication & 70 & 23 \\
abuse product & misuse product & 39 & 58 \\
attack problem & address problem & 1,707 & 35,477 \\
attack issue & address issue & 239 & 79,175 \\
attack need & address need & 20 & 20,487 \\
attack question & address question & 114 & 21,778 \\
attack challenge & address challenge & 83 & 11,541 \\
attack point & address point & 232 & 4,098 \\
attack situation & address situation & 45 & 3,465 \\
attack topic & address topic & 22 & 4,066 \\
attack change & address change & 54 & 3,579 \\
attack cause & address cause & 155 & 3,094 \\
attack matter & address matter & 49 & 3,451 \\
boost economy & improve economy & 4,687 & 2,755 \\
boost service & improve service & 177 & 16,003 \\
boost system & improve system & 1,829 & 6,837 \\
boost situation & improve situation & 16 & 6,316 \\
boost process & improve process & 98 & 5,235 \\
boost education & improve education & 87 & 3,837 \\
boost work & improve work & 80 & 3,424 \\
boost business & improve business & 871 & 1,910 \\
boost result & improve result & 211 & 2,630 \\
boost number & improve number & 1,749 & 710 \\
break agreement & end agreement & 1,225 & 329 \\
break cycle & end cycle & 4,193 & 647 \\
break relationship & end relationship & 1,318 & 2,321 \\
break contract & end contract & 1,571 & 590 \\
break marriage & end marriage & 694 & 1,094 \\
break process & end process & 945 & 523 \\
break pattern & end pattern & 1,060 & 50 \\
breathe life & instill life & 6,227 & 30 \\
breathe sense & instill sense & 25 & 1,074 \\
breathe confidence & instill confidence & 30 & 1,144 \\
breathe value & instill value & 26 & 659 \\
breathe spirit & instill spirit & 391 & 128 \\
breathe love & instill love & 66 & 308 \\
breathe hope & instill hope & 47 & 133 \\
breathe idea & instill idea & 16 & 134 \\
breathe passion & instill passion & 16 & 96 \\
buy story & believe story & 724 & 1,823 \\
buy word & believe word & 125 & 2,997 \\
buy lie & believe lie & 260 & 1,805 \\
cast doubt & cause doubt & 8,494 & 257 \\
cast issue & cause issue & 98 & 6,530 \\
cast fear & cause fear & 282 & 1,365 \\
catch disease & get disease & 1,201 & 2,662 \\
catch idea & get idea & 201 & 52,090 \\
catch chance & get chance & 67 & 55,978 \\
catch information & get information & 72 & 36,240 \\
catch result & get result & 47 & 36,573 \\
catch message & get message & 134 & 27,705 \\
catch point & get point & 147 & 23,962 \\
catch call & get call & 43 & 20,692 \\
catch problem & get problem & 594 & 14,694 \\
catch opportunity & get opportunity & 54 & 13,177 \\
close investigation & end investigation & 504 & 124 \\
close season & end season & 1,037 & 2,805 \\
close deal & end deal & 2,744 & 168 \\
close case & end case & 3,047 & 240 \\
close debate & end debate & 601 & 762 \\
close operation & end operation & 643 & 309 \\
close story & end story & 122 & 903 \\
close deal & finalise deal & 2,744 & 424 \\
close case & finalise case & 3,047 & 66 \\
close plan & finalise plan & 98 & 825 \\
close arrangement & finalise arrangement & 27 & 367 \\
close agreement & finalise agreement & 65 & 293 \\
close project & finalise project & 381 & 64 \\
close deal & finalize deal & 2,744 & 468 \\
close case & finalize case & 3,047 & 24 \\
close plan & finalize plan & 98 & 867 \\
close arrangement & finalize arrangement & 27 & 156 \\
close agreement & finalize agreement & 65 & 438 \\
close project & finalize project & 381 & 86 \\
cloud memory & impair memory & 77 & 297 \\
cloud ability & impair ability & 39 & 1,977 \\
cloud judgement & impair judgement & 778 & 244 \\
cloud judgment & impair judgment & 702 & 376 \\
cloud mind & impair mind & 534 & 30 \\
cloud vision & impair vision & 335 & 424 \\
cloud thinking & impair thinking & 114 & 48 \\
cloud perception & impair perception & 85 & 43 \\
cloud understanding & impair understanding & 70 & 27 \\
color judgement & affect judgement & 12 & 301 \\
color decision & affect decision & 36 & 3,236 \\
color choice & affect choice & 10 & 1,555 \\
color perception & affect perception & 145 & 1,279 \\
color experience & affect experience & 77 & 1,222 \\
color interpretation & affect interpretation & 38 & 498 \\
colour judgement & affect judgement & 69 & 301 \\
colour decision & affect decision & 32 & 3,236 \\
colour choice & affect choice & 22 & 1,555 \\
colour perception & affect perception & 135 & 1,279 \\
colour experience & affect experience & 55 & 1,222 \\
colour interpretation & affect interpretation & 29 & 498 \\
deflate economy & reduce economy & 71 & 181 \\
deflate cost & reduce cost & 12 & 51,650 \\
deflate price & reduce price & 63 & 6,766 \\
deflate value & reduce value & 58 & 4,430 \\
deflate supply & reduce supply & 12 & 1,784 \\
deflate wage & reduce wage & 23 & 1,130 \\
deflate market & reduce market & 15 & 199 \\
deflate currency & reduce currency & 24 & 26 \\
devour book & read book & 587 & 104,728 \\
devour article & read article & 22 & 55,121 \\
devour story & read story & 23 & 17,990 \\
devour page & read page & 37 & 10,667 \\
devour novel & read novel & 77 & 8,517 \\
devour information & read information & 58 & 6,787 \\
devour chapter & read chapter & 21 & 6,345 \\
devour news & read news & 15 & 4,480 \\
digest information & comprehend information & 496 & 124 \\
digest meaning & comprehend meaning & 18 & 352 \\
digest material & comprehend material & 192 & 113 \\
digest fact & comprehend fact & 83 & 139 \\
digest text & comprehend text & 21 & 217 \\
digest concept & comprehend concept & 30 & 190 \\
digest idea & comprehend idea & 60 & 172 \\
digest word & comprehend word & 52 & 174 \\
digest content & comprehend content & 154 & 49 \\
digest situation & comprehend situation & 12 & 115 \\
digest message & comprehend message & 49 & 87 \\
disown past & reject past & 17 & 67 \\
disown idea & reject idea & 14 & 7,519 \\
disown policy & reject policy & 9 & 567 \\
disown responsibility & reject responsibility & 37 & 125 \\
drop price & reduce price & 2,439 & 6,766 \\
drop cost & reduce cost & 218 & 51,650 \\
drop rate & reduce rate & 817 & 15,826 \\
drop temperature & reduce temperature & 357 & 2,077 \\
drown trouble & forget trouble & 9 & 268 \\
drown pain & forget pain & 29 & 414 \\
drown problem & forget problem & 16 & 423 \\
drown feeling & forget feeling & 16 & 373 \\
dull appetite & decrease appetite & 21 & 240 \\
dull pain & decrease pain & 315 & 384 \\
dull sense & decrease sense & 316 & 54 \\
dull noise & decrease noise & 11 & 72 \\
dull feeling & decrease feeling & 23 & 58 \\
find excuse & make excuse & 2,374 & 9,643 \\
find way & make way & 149,262 & 42,972 \\
find connection & make connection & 3,155 & 29,720 \\
float idea & suggest idea & 1,798 & 3,386 \\
float theory & suggest theory & 108 & 383 \\
float concept & suggest concept & 48 & 316 \\
flood market & saturate market & 2,989 & 1,167 \\
follow profession & practice profession & 228 & 610 \\
follow religion & practice religion & 1,371 & 2,125 \\
follow activity & practice activity & 824 & 230 \\
follow profession & practise profession & 228 & 266 \\
follow religion & practise religion & 1,371 & 537 \\
follow activity & practise activity & 824 & 72 \\
frame question & pose question & 1,553 & 18,561 \\
frame problem & pose problem & 570 & 12,804 \\
frame challenge & pose challenge & 97 & 9,544 \\
frame issue & pose issue & 1,574 & 945 \\
frame debate & pose debate & 1,011 & 17 \\
frame concern & pose concern & 49 & 588 \\
frame argument & pose argument & 535 & 117 \\
frame idea & pose idea & 147 & 120 \\
frame hypothesis & pose hypothesis & 102 & 56 \\
fuel debate & stimulate debate & 595 & 1,351 \\
fuel growth & stimulate growth & 1,929 & 4,940 \\
fuel economy & stimulate economy & 733 & 4,359 \\
fuel interest & stimulate interest & 437 & 2,354 \\
fuel discussion & stimulate discussion & 149 & 1,821 \\
fuel demand & stimulate demand & 553 & 1,262 \\
fuel activity & stimulate activity & 154 & 1,487 \\
fuel imagination & stimulate imagination & 214 & 775 \\
fuel creativity & stimulate creativity & 112 & 473 \\
fuel conversation & stimulate conversation & 57 & 385 \\
grasp meaning & understand meaning & 802 & 5,927 \\
grasp concept & understand concept & 2,149 & 9,262 \\
grasp issue & understand issue & 284 & 8,076 \\
grasp point & understand point & 531 & 6,359 \\
grasp problem & understand problem & 188 & 6,464 \\
grasp situation & understand situation & 220 & 3,935 \\
grasp reason & understand reason & 59 & 4,449 \\
grasp language & understand language & 87 & 4,220 \\
grasp idea & understand idea & 913 & 2,688 \\
grasp risk & understand risk & 23 & 3,262 \\
grasp question & understand question & 30 & 3,016 \\
juggle job & manage job & 353 & 425 \\
juggle project & manage project & 149 & 7,242 \\
juggle work & manage work & 389 & 1,291 \\
juggle life & manage life & 234 & 1,136 \\
juggle career & manage career & 168 & 578 \\
juggle school & manage school & 38 & 825 \\
kill proposal & cancel proposal & 156 & 24 \\
kill project & cancel project & 517 & 1,284 \\
kill bill & cancel bill & 1043 & 59 \\
kill program & cancel program & 403 & 934 \\
kill process & cancel process & 813 & 124 \\
kill agreement & cancel agreement & 20 & 717 \\
kill deal & cancel deal & 253 & 279 \\
leak report & disclose report & 330 & 280 \\
leak information & disclose information & 2,977 & 13,122 \\
leak document & disclose document & 884 & 603 \\
leak story & disclose story & 361 & 56 \\
mount production & organise production & 322 & 305 \\
mount event & organise event & 49 & 14,000 \\
mount campaign & organise campaign & 2,288 & 1,007 \\
mount conference & organise conference & 19 & 3,384 \\
mount exhibition & organise exhibition & 765 & 885 \\
mount demonstration & organise demonstration & 96 & 1,286 \\
mount protest & organise protest & 237 & 1,455 \\
mount production & organize production & 322 & 316 \\
mount event & organize event & 49 & 5,955 \\
mount campaign & organize campaign & 2,288 & 1,197 \\
mount conference & organize conference & 19 & 2,947 \\
mount exhibition & organize exhibition & 765 & 1,229 \\
mount demonstration & organize demonstration & 96 & 1,050 \\
mount protest & organize protest & 237 & 1,552 \\
poison mind & corrupt mind & 596 & 402 \\
poison system & corrupt system & 148 & 556 \\
poison process & corrupt process & 31 & 343 \\
poison soul & corrupt soul & 88 & 188 \\
poison relationship & corrupt relationship & 141 & 48 \\
pour money & invest money & 2,383 & 10,003 \\
pour fortune & invest fortune & 20 & 134 \\
push drug & sell drug & 311 & 3,680 \\
recapture feeling & recall feeling & 63 & 245 \\
recapture memory & recall memory & 34 & 1,112 \\
recapture moment & recall moment & 55 & 961 \\
recapture experience & recall experience & 43 & 887 \\
shake confidence & damage confidence & 870 & 440 \\
shake foundation & damage foundation & 1,124 & 138 \\
shape result & determine result & 100 & 1,125 \\
shape life & determine life & 3,737 & 848 \\
shape outcome & determine outcome & 451 & 3,472 \\
shape success & determine success & 104 & 2,795 \\
shape strategy & determine strategy & 724 & 1,061 \\
shipwreck career & ruin career & 2 & 1,165 \\
sow doubt & cause doubt & 246 & 257 \\
sow death & cause death & 44 & 18,023 \\
sow confusion & cause confusion & 345 & 7,077 \\
sow chaos & cause chaos & 76 & 2,622 \\
sow conflict & cause conflict & 18 & 2,110 \\
sow panic & cause panic & 33 & 1,808 \\
sow fear & cause fear & 133 & 1,365 \\
sow violence & cause violence & 22 & 930 \\
sow uncertainty & cause uncertainty & 11 & 591 \\
sow terror & cause terror & 85 & 252 \\
sow hatred & cause hatred & 68 & 181 \\
stir excitement & cause excitement & 113 & 766 \\
stir confusion & cause confusion & 22 & 7,077 \\
stir reaction & cause reaction & 94 & 3,894 \\
stir feeling & cause feeling & 530 & 1,357 \\
stir emotion & cause emotion & 1,009 & 227 \\
suck worker & attract worker & 12 & 690 \\
suck talent & attract talent & 31 & 1,693 \\
tackle question & address question & 2,250 & 21,778 \\
tackle issue & address issue & 11,995 & 79,175 \\
tackle problem & address problem & 15,180 & 35,477 \\
tackle concern & address concern & 299 & 17,506 \\
tackle challenge & address challenge & 3,352 & 11,541 \\
tackle situation & address situation & 533 & 3,465 \\
tackle point & address point & 111 & 4,098 \\
tackle crisis & address crisis & 1,322 & 2,491 \\
tackle matter & address matter & 347 & 3,451 \\
tackle inequality & address inequality & 1,445 & 1543 \\
tackle task & address task & 1,017 & 396 \\
taste freedom & experience freedom & 145 & 594 \\
taste pain & experience pain & 21 & 5,984 \\
taste life & experience life & 130 & 3,837 \\
taste joy & experience joy & 102 & 2,071 \\
throw remark & make remark & 72 & 13,467 \\
throw comment & make comment & 282 & 47,768 \\
twist word & misinterpret word & 876 & 187 \\
twist fact & misinterpret fact & 599 & 47 \\
twist meaning & misinterpret meaning & 218 & 109 \\
twist comment & misinterpret comment & 37 & 148 \\
twist situation & misinterpret situation & 39 & 72 \\
twist information & misinterpret information & 28 & 114 \\
twist message & misinterpret message & 46 & 101 \\
wear smile & have smile & 876 & 5,975 \\

\end{longtable}
\label{tab:app:met-lit-count1}
\twocolumn

\end{document}